\documentclass[letterpaper]{article} 
\usepackage{aaai2026}
\usepackage{times}  
\usepackage{helvet}  
\usepackage{courier}  
\usepackage[hyphens]{url}  
\usepackage{graphicx} 
\usepackage{natbib}  
\usepackage{caption} 
\usepackage{algorithm}
\usepackage{algorithmic}
\usepackage{url}  
\usepackage{booktabs}
\usepackage{amsfonts}  
\usepackage{nicefrac} 
\usepackage{microtype}   
\usepackage{fancyhdr}      
\usepackage{bm}
\usepackage{amsmath}
\usepackage{algorithmic}
\usepackage{algorithm}
\usepackage{multirow}
\usepackage{graphicx}
\usepackage{subfigure}
\usepackage{amsthm}
\usepackage{multicol}
\usepackage{pdfpages}
\usepackage{makecell}
\usepackage{xspace}
\usepackage{float}
\usepackage{threeparttable}
\usepackage{soul}
\usepackage{array}
\usepackage{amsthm,amsmath,amssymb}
\usepackage{dsfont}
\usepackage{pifont}
\usepackage{amssymb}
\usepackage{caption}
\usepackage{graphicx, subfig}
\usepackage{xcolor}

\makeatletter
\DeclareRobustCommand\onedot{\futurelet\@let@token\@onedot}
\def\@onedot{\ifx\@let@token.\else.\null\fi\xspace}
\def\eg{\emph{e.g}\onedot}
\def\ie{\emph{i.e}\onedot}
\def\etc{\emph{etc}\onedot}

\def\etal{\emph{et al}\onedot}
\makeatother
\usepackage{newfloat}
\usepackage{listings}
\DeclareCaptionStyle{ruled}{labelfont=normalfont,labelsep=colon,strut=off} 
\floatstyle{ruled}
\newfloat{listing}{tb}{lst}{}
\floatname{listing}{Listing}
\title{CLASP: Cross-modal Salient Anchor-based Semantic Propagation for \\ Weakly-supervised Dense Audio-Visual Event Localization}
\author{
    Jinxing Zhou\textsuperscript{\rm 1}\equalcontrib,
    Ziheng Zhou\textsuperscript{\rm 2}\equalcontrib,
    Yanghao Zhou\textsuperscript{\rm 4},
    Yuxin Mao\textsuperscript{\rm 5}, 
    Zhangling Duan\textsuperscript{\rm 3},
    Dan Guo\textsuperscript{\rm 2,3}\thanks{Corresponding author}
}

\affiliations{
    \textsuperscript{\rm 1}Mohamed Bin Zayed University of Artificial Intelligence~~~~
    \textsuperscript{\rm 2}Hefei University of Technology\\
    \textsuperscript{\rm 3}Hefei Comprehensive National Science Center~~~~
     \textsuperscript{\rm 4}National University of Singapore~~~~
     \textsuperscript{\rm 5}OpenNLP Lab\\
     }
\usepackage{bibentry}

\begin{document}

\maketitle

\begin{abstract}
The Dense Audio-Visual Event Localization (DAVEL) task aims to temporally localize events in untrimmed videos that occur simultaneously in both the audio and visual modalities.
This paper explores DAVEL under a new and more challenging weakly-supervised setting (W-DAVEL task), where only video-level event labels are provided and the temporal boundaries of each event are unknown.
We address W-DAVEL by exploiting \textit{cross-modal salient anchors}, which are defined as reliable timestamps that are well predicted under weak supervision and exhibit highly consistent event semantics across audio and visual modalities.
Specifically, we propose a \textit{Mutual Event Agreement Evaluation} module, which generates an agreement score by measuring the discrepancy between the predicted audio and visual event classes.
Then, the agreement score is utilized in a \textit{Cross-modal Salient Anchor Identification} module, which identifies the audio and visual anchor features through global-video and local temporal window identification mechanisms.
The anchor features after multimodal integration are fed into an \textit{Anchor-based Temporal Propagation} module to enhance event semantic encoding in the original temporal audio and visual features, facilitating better temporal localization under weak supervision.
We establish benchmarks for W-DAVEL on both the UnAV-100 and ActivityNet1.3 datasets. Extensive experiments demonstrate that our method achieves state-of-the-art performance.

\end{abstract}


\section{Introduction}
In recent years, audio-visual learning~\cite{shen2023fine,zhou2024label,zhou2025mettle,liu2025towards,li2025patch,li2024object,mao2024tavgbench,zhoualoha,guo2024enhance,zhao2025multimodal} has received increasing attention due to its potential to jointly model and reason over auditory and visual modalities, facilitating a more comprehensive understanding of complex real-world environments.
This line of research has made significant progress and plays a crucial role in advancing multimodal artificial intelligence.
Various audiovisual learning tasks have been widely explored, such as audio-visual event perception~\cite{tian2018audio,tian2020unified,zhou2024towards}, sound source localization~\cite{arandjelovic2017look}, audio-visual segmentation~\cite{zhou2022avs,zhou2023avss,guo2023audio}, and more.
However, the majority of existing works focus on short-term audio-visual scene understanding.

\begin{figure}[t]
  \centering
  \includegraphics[width=\linewidth]{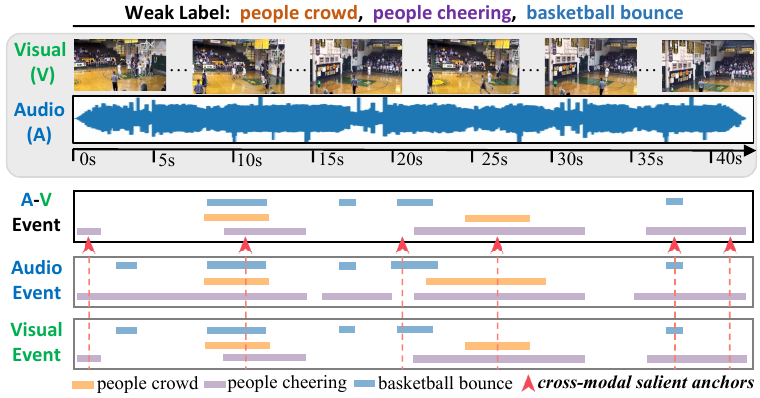}
  \caption{\textbf{Illustration of the Weakly-supervised Dense Audio-Visual Event Localization (W-DAVEL) task.} 
    The W-DAVEL task aims to temporally localize events that co-occur in both audio and visual modalities, using only weak supervision from video-level event labels.
    }
  \vspace{-2ex}
  \label{fig:intro}
\end{figure}

Recently, Geng~\etal~\cite{geng2023dense} proposed the \textit{Dense Audio-Visual Event Localization (DAVEL)} task, aiming to understand long, untrimmed videos.
As shown in Fig.~\ref{fig:intro}, given a long video with its audio track, the DAVEL task aims to temporally localize events occurring in both audio and visual modalities, \ie, the intersection set of the audio events and visual events.
Each video usually contains multiple events, and each event may occur at multiple temporal positions, which makes DAVEL a highly challenging task.
To date, several studies have attempted to address this problem, with most focusing on more effective audio-visual fusion~\cite{geng2023dense,xing2024loco,zhou2025dense}.
Although significant progress has been made by these approaches, they are all developed under a \textit{fully-supervised} setting, where the specific start and end timestamp labels of each audio-visual event are provided.
However, manually obtaining such fine-grained annotations is time-consuming, labor-intensive, and difficult to scale to large datasets.
Therefore, the study of DAVEL under weak supervision emerges as an urgent and practical direction.

In this paper, we introduce the \textbf{Weakly-supervised Dense Audio-Visual Event Localization (W-DAVEL)} task, which aims to temporally localize audio-visual events using \textit{only video-level labels as supervision}.
As shown in Fig.~\ref{fig:intro}, {given the weak label of `{people crowd, people cheering, basketball bounce}'}, the model needs to predict not only the event categories but also their corresponding temporal boundaries.
In other words, for each timestamp, the model is expected to predict the audio-visual event categories.
In the W-DAVEL task, the weak label can provide explicit supervision indicating the presence of events in the video.
However, a critical challenge is how to handle the absence of category supervision along the temporal timeline.
We address this challenge by distributing the event semantics, derived from reliable timestamps guided by video-level supervision, across the temporal dimension through a learnable network.
Specifically, even with only video-level event labels, there exist temporal timestamps where the model can confidently infer the audio-visual event categories.
We define such timestamps as \textit{\textbf{cross-modal salient anchors}}, which contain the most consistent event semantics across both audio and visual modalities and are distinguishable from the remaining temporal segments.
We then model the relationships between temporal features and these anchor features to determine event semantics across the entire timeline.

Based on this idea, we propose a \textit{\textbf{C}ross-modal sa\textbf{l}ient \textbf{A}nchor-based \textbf{S}emantic \textbf{P}ropagation (\textbf{CLASP})} method.
Specifically, to identify the salient anchors, we first introduce a \textit{\textbf{Mutual Event Agreement Evaluation (MEAE)}} module.
Modality-specific classifiers independently predict audio and visual event probabilities.
MEAE measures the discrepancy between the predicted audio and visual event probabilities using the Jensen-Shannon Divergence, from which a mutual event agreement score is obtained.
This score reflects the model's confidence in the consistency of audiovisual predictions at each timestamp.
Based on this agreement score, we design a \textit{\textbf{Cross-modal Salient Anchor Identification (CSAI)}} module, which identifies the anchor timestamps through two mechanisms:
Global Anchor Identification selects anchors with the top-$K$ agreement scores from a global comparison across the entire video, while Local Anchor Identification follows a similar process but is restricted to multiple local temporal windows.
CSAI is applied to both audio and visual modalities. 
The identified audio and visual anchor features are fused into compact multimodal anchor features, which encode explicit audio-visual event semantics.
Then, an \textit{\textbf{Anchor-based Temporal Propagation (ATP)}} module leverages the anchor features to enhance temporal audio and visual features, propagating event semantics from anchors to fine-grained temporal timestamps, facilitating event classification along timeline.

We evaluate our method and establish benchmarks for the W-DAVEL task through comparisons with prior methods from related tasks on the UnAV-100~\cite{geng2023dense} and ActivityNet1.3~\cite{heilbron2015activitynet} datasets.
Experimental results demonstrate that our method outperforms existing approaches by a large margin.
For instance, our method surpasses the previous state-of-the-art (SOTA) method CCNet~\cite{zhou2025dense} by 3.1\% mAP on UnAV-100 dataset.

In summary, our main contributions are:
\begin{itemize}
\item We explore the practical and challenging W-DAVEL task, which aims to perform dense audio-visual event localization using only video-level event labels as supervision.
\item We propose a salient anchor-based CLASP method, which consists of three core modules: Mutual Event Agreement Evaluation, Cross-modal Salient Anchor Identification, and Anchor-based Temporal Propagation.
\item We establish benchmarks for the W-DAVEL task on the UnAV-100 and ActivityNet1.3 datasets, achieving new state-of-the-art results compared to previous methods.
\end{itemize}

\section{Related Work}
\noindent\textbf{Audio-Visual Event Perception (AVEP)} aims to temporally localize events in specific categories that occur in audible videos. 
According to the types of the target events, existing research can be categorized into two branches: Audio-Visual Event Localization (AVEL)~\cite{tian2018audio} and Audio-Visual Video Parsing (AVVP)~\cite{tian2020unified}.
The AVEL task focuses on identifying the audio-visual events that occur simultaneously in both modalities, while the AVVP task requires further distinguishing events occurring in each individual modality, \ie, the audio events and visual events. 
To date, many methods for these tasks can operate under weakly supervised settings, where only the video-level event labels are provided.
From a model architecture perspective, mainstream methods emphasize audiovisual interaction through various cross-attention fusion strategies~\cite {tian2018audio,xu2020cross,ramaswamy2020makes,mahmud2023ave,yu2022mmpyr,mo2022multi,jiang2022dhhn}, positive segment discovery~\cite{zhou2021positive,zhou2022cpsp,gao2023collecting}, and background suppression~\cite{xia2022cmbs,he2024cace}.
To improve the weak supervision, some methods aim to obtain modality-specific weak labels by denoising the available video label~\cite{wu2021exploring,cheng2022joint}, or generate temporally specific pseudo-labels~\cite{lai2023modality,zhou2023improving,zhou2024advancing,fan2023revisit} by utilizing external large foundation models, such as CLIP~\cite{radford2021CLIP} and CLAP~\cite{wu2023clap}.
Although these prior works have made notable progress in addressing weakly supervised AVEP tasks, they are limited to trimmed video scenarios. 
This study aims to fill the gap in audio-visual event perception for long, untrimmed videos under weak supervision.

\noindent\textbf{Dense Audio-Visual Event Localization (DAVEL)} aims to densely identify and localize audio-visual events in untrimmed videos.
The pioneering work~\cite{geng2023dense} proposes a strong baseline alongside the UnAV-100 benchmark. A cross-modal pyramid transformer encoder is employed to fuse audiovisual features at various temporal scales, and a temporal dependency modeling module is designed to capture correlations between multiple event categories. 
Subsequently, Xing~\etal~\cite{xing2024loco} leverage the local temporal continuity of audio-visual events to filter out irrelevant cross-modal signals and enhance cross-modal fusion via an adaptive attention mechanism.
Zhou~\etal~\cite{zhou2025dense} introduce CCNet, which improves audiovisual feature fusion via cross-modal consistency and collaborative modeling across multiple temporal granularities.
These methods are developed under the fully supervised setting, which requires fine-grained temporal labels for effective model training.
{In addition, several approaches, specifically designed for short-term AVEP tasks, such as PPAE~\cite{gao2025learning} and NREP~\cite{jiang2024resisting}, also evaluate their models on the UnAV-100 dataset, yet their performances on long-term event localization are far from satisfactory.}
In contrast, we provide a formal analysis of the weakly supervised DAVEL task and address it by exploring cross-modal salient anchors, achieving superior temporal localization results.

\begin{figure*}[t]
    \centering
  \includegraphics[width=\linewidth]{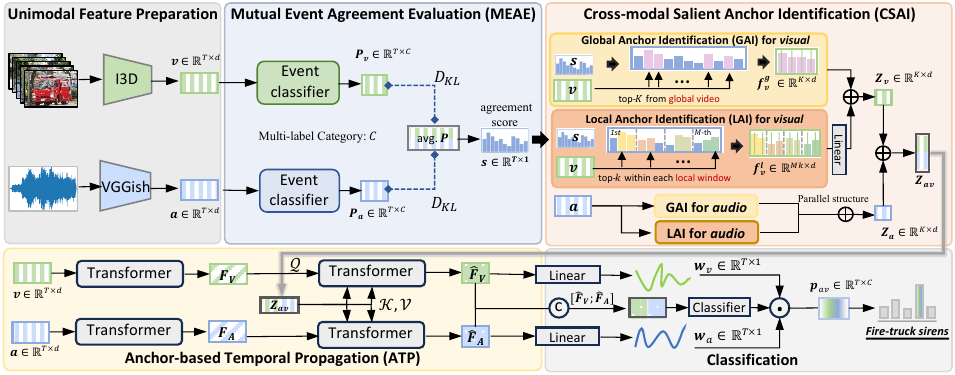}
    \caption{\textbf{Illustration of our cross-modal salient anchor-based semantic propagation (CLASP) framework for W-DAVEL.}}
    \label{fig:framework}
\end{figure*}

\section{Methodology}
\subsection{Problem Statement}
Given a video with $T$ segments at equal intervals, let the $t$-th visual and audio segments be denoted as $V_t$ and $A_t$ ($t=\{1,...,T\}$), respectively.
During the training process, only the video-level event label $\bm{y}= \{y_i\}_{i=1}^C \in \mathbb{R}^{1\times C}$ is provided. Here, $C$ is the total number of classes of audio-visual events, and $y_i$ takes a value from $\{0, 1\}$, where $y_i = 1$ indicates that the $i$-th category exists in the video (but the specific timestamp is unknown).
During the evaluation phase, the segment-level ground truth $Y = \{y_n = (t^s_n, t^e_n, c_n) \}_{n=1}^N$ is available for evaluation, where $N$ is the total number of events occurring in the video, $t^s_n$ and $t^e_n$ are the start and end timestamps of the $n$-th audio-visual event, respectively, and $c_n$ denotes the corresponding event category.

Therefore, a W-DAVEL model aims to predict the start and end timestamps for each identified event.
Prior methods~\cite{geng2023dense,zhou2025dense} for fully supervised DAVEL typically formulate the task as a joint category classification and timestamp regression problem.
However, the regression subproblem is challenging to solve for the W-DAVEL task due to the lack of explicit timestamp supervision. 
Therefore, we formulate the W-DAVEL task as an explicit event classification problem, in which the model estimates event category probabilities $\bm{p}_{av} \in \mathbb{R}^{T \times C}$.


{\subsection{Framework Overview}}
An overview of the proposed framework is illustrated in Fig.~\ref{fig:framework}. It consists of five main modules.
(1) In the \textit{\textbf{Unimodal Feature Preparation}}, modality-specific pretrained backbones are used to extract preliminary features for both audio and visual modalities.
(2) Based on the extracted features, the \textit{\textbf{Mutual Event Agreement Evaluation (MEAE)}} module measures the semantic consistency of event predictions between the audio and visual modalities. An event agreement score is obtained by comparing the predicted audio and visual event probabilities.
(3) Given the event agreement score, the \textit{\textbf{Cross-modal Salient Anchor Identification (CSAI)}} module employs two mechanisms to identify salient anchors along the temporal axis.
The salient anchors refer to timestamps with the most explicit and consistent event categories across audio and visual modalities, which correspond to large values in the event agreement score.
Specifically, Global Anchor Identification directly selects anchors from the entire video, while Local Anchor Identification performs fine-grained anchor selection within multiple, local temporal windows. 
The anchor audio and visual features identified by both mechanisms are fused to form compact multimodal anchor representations.
(4) These fused features subsequently enhance the intra-modal temporal audio and visual features via the \textit{\textbf{Anchor-based Temporal Propagation (ATP)}}.
(5) Finally, the updated audio and visual features are sent to the \textbf{\textit{Classification}} module for event predictions.
We elaborate on each module in the next subsections.

\subsection{Unimodal Feature Preparation}
Given the audio and visual segments $\{A_t, V_t\}_{t=1}^T$, we follow prior work~\cite{geng2023dense} to extract initial features.
Specifically, VGGish~\cite{hershey2017vggish}, pretrained on the AudioSet~\cite{gemmeke2017audioset} dataset, is employed to extract audio features, denoted as $\bm{a} \in \mathbb{R}^{T \times d}$, where $d$ is the channel dimension.
The two-stream I3D~\cite{carreira2017quo}, pretrained on Kinetics~\cite{kay2017kinetics}, is employed to extract visual features, denoted as $\bm{v} \in \mathbb{R}^{T \times d}$.
Here, the audio and visual features are projected into a shared feature space using two convolutional layers with ReLU.

\subsection{Mutual Event Agreement Evaluation}
Our method handles W-DAVEL task by highlighting cross-modal salient anchors, \ie, timestamps that exhibit shared salient event semantics across both audio and visual modalities.
To this end, we propose to first measure the mutual agreement on event categories along the temporal dimension.

Given the audio and visual features $\bm{a} \in \mathbb{R}^{T \times d}$ and $\bm{v} \in \mathbb{R}^{T \times d}$, we first predict event categories independently for each modality using separate classifiers. 
This is implemented using two linear layers with a LeakyReLU activation in between, formulated as:
\begin{equation}
\begin{gathered}
        \bm{P}_a = \sigma(  (\bm{a} \bm{W}_1^a )\bm{W}_2^a) , ~~
    \bm{P}_v = \sigma( (\bm{v} \bm{W}_1^v )\bm{W}_2^v),
    \label{eq:Pa_Pv}
\end{gathered}
\end{equation}
where $\bm{W}^m_1 \in \mathbb{R}^{d\times d}$, $\bm{W}_2^m \in \mathbb{R}^{d \times C}$ ($m \in \{a, v\}$), and $\sigma$ is the sigmoid function to enable multi-label event prediction.

$\bm{P}_a, \bm{P}_v \in \mathbb{R}^{T \times C}$ represent the event probabilities at each timestamp.
A target audio-visual event is considered present only when the predicted class probability is high in both modalities, \ie, both modalities agree that the event exists.
We first measure the discrepancy between the predicted distributions $\bm{P}_a$ and $\bm{P}_v$ using the tool of Jensen-Shannon Divergence~\cite{fuglede2004jensen}, as follows:
\begin{equation}
\begin{gathered}
    \overline{\bm{P}} = \frac{1}{2}(\bm{P}_a + \bm{P}_v),\\
    \bm{d}_{JSD} = \frac{1}{2}{D_{KL}}(\overline{\bm{P}}, \bm{P}_a) + \frac{1}{2}{D_{KL}}(\overline{\bm{P}}, \bm{P}_v), \\
\end{gathered}
\end{equation}
where $D_{KL}$ is the Kullback-Leibler Divergence. $\bm{d}_{JSD} \in \mathbb{R}^{T \times 1}$.
Accordingly, the audiovisual mutual agreement score $\bm{s}$ can be obtained by:
\begin{equation}
   \bm{s} = 1 - \bm{d}_{JSD},
\end{equation}
where $\bm{s} \in \mathbb{R}^{T \times 1}$.
The higher the value of $\bm{d}_{JSD}$, the more different the distributions of $\bm{P}_a$ and $\bm{P}_v$ are, indicating lower overlap in event categories, resulting in a lower mutual event agreement score.

\subsection{Cross-modal Salient Anchor Identification}
After obtaining the mutual agreement score $\bm{s} \in \mathbb{R}^{T \times 1}$, we employ it as an indicator to identify cross-modal salient anchors along the temporal axis.
We develop two mechanisms:

\noindent\textbf{Global Anchor Identification (GAI).}
The GAI aims to identify the salient anchors by performing a global comparison over the entire video timeline.
Specifically, timestamps with the top-$K$ values in $\bm{s}$ are selected.
Let $\mathcal{I}^g$ denote the index of the selected timestamps, where each index $\mathcal{I}^g_i$ belongs to $\{1, ..., T\}$ for $i = 1, ..., K$.
Using $\mathcal{I}^g$, we extract the corresponding audio and visual features from $\bm{a}$ and $\bm{v}$, respectively, denoted as $\bm{f}_a^g \in \mathbb{R}^{K \times d}$ and $\bm{f}_v^g \in \mathbb{R}^{K \times d}$.
$\bm{f}_a^g$ and $\bm{f}_v^g$ represent the features that exhibit the most consistent event semantics across both audio and visual modalities.


\noindent\textbf{Local Anchor Identification (LAI).}
In a long video, audio-visual events often appear at multiple temporal positions.
In addition to GAI, we propose LAI to identify salient anchors through a finer-grained temporal comparison.
Specifically, {as shown in Fig.~\ref{fig:framework}}, the video is divided into $M$ temporal windows, each containing $\lfloor \frac{T}{M} \rfloor$ segments. 
For each temporal window, we extract the corresponding slice of the mutual agreement score $\bm{s}$, denoted as $\bm{s}_m$ ($m=1,...,M$).
Then, operations similar to GAI are applied to identify the anchors with the top-$k$ values of $\bm{s}_m$.
We denote the obtained audio and visual anchor features as $ \bm{f}_a^{l,m}$ and $ \bm{f}_v^{l,m}$, where $\bm{f}^{l,m} \in \mathbb{R}^{k \times d}$ is the anchor feature set for the $m$-th temporal window.
Therefore, the anchor features across all $M$ windows are aggregated and denoted as $\bm{f}_a^l = \{ \bm{f}_a^{l,m} \}_{m=1}^M $ and $\bm{f}_v^l = \{ \bm{f}_v^{l,m} \}_{m=1}^M \in \mathbb{R}^{Mk \times d}$.
A linear layer is then applied to transform $\bm{f}_a^l$ and $\bm{f}_v^l$, aligning them with the $K$ anchors obtained from GAI, yielding $\bm{f}_a^l, \bm{f}_v^l \in \mathbb{R}^{K \times d}$.

After obtaining the anchor features from both the global video level and the local-temporal window level, we aggregate the two types of anchor features as follows:
\begin{equation}
    \begin{gathered}
        \bm{Z}_a = \bm{f}_a^g + \bm{f}_a^l,~~ 
        \bm{Z}_v = \bm{f}_v^g + \bm{f}_v^l, \\
        \bm{Z}_{av} =([\bm{Z}_a; \bm{Z}_v]\bm{W}_3) \bm{W}_4,
    \end{gathered}
    \label{eq:anchor_a_v}
\end{equation}
{where $\bm{W}_3 \in \mathbb{R}^{2d \times 2d}$, $\bm{W}_4 \in \mathbb{R}^{2d \times d}$}. $[\cdot; \cdot]$ represents the feature concatenation along the channel dimension. $\bm{Z}_{av} \in \mathbb{R}^{K \times d}$ is the ultimate anchor features, which contain the salient event category semantics (events mostly likely to occur) in both audio and visual modalities.

\subsection{Anchor-based Temporal Propagation}
The W-DAVEL task requires identifying the event category for each timestamp.
Therefore, it is important to capture \textit{temporal} features that accurately reflect event semantics.
Our Anchor-based Temporal Propagation (ATP) module enhances temporal interactions using the identified anchors. 
First, the original audio $\bm{a}$ and visual $\bm{v}$ features are passed through separate unimodal one-layer Transformers~\cite{vaswani2017attention} to encode the temporal relations within each modality, yielding the updated features $\bm{F}_A, \bm{F}_V \in \mathbb{R}^{T\times d}$.

Then, we model the interaction between the modality-specific features ($\bm{F}_A / \bm{F}_V$) and the multimodal anchor feature $\bm{Z}_{av}$ via a cross-attention mechanism, formulated as:
\begin{equation}
    \begin{gathered}
        \bm{\hat{F}}_A = \bm{F}_A + \text{MHA}(\bm{F}_A, \bm{Z}_{av}, \bm{Z}_{av}), \\
        \bm{\hat{F}}_V = \bm{F}_V + \text{MHA}(\bm{F}_V, \bm{Z}_{av}, \bm{Z}_{av}), \\
  \text{MHA}(\bm{\mathcal{Q}}, \bm{\mathcal{K}}, \bm{\mathcal{V}})  = \delta\left( \frac{\bm{\mathcal{Q}\bm{W}}^Q (\bm{\mathcal{K}}\bm{W}^K )^{\top}} {\sqrt{d}} \right) \bm{\mathcal{V}}\bm{W}^V,
    \end{gathered}
    \label{eq:ATP}
\end{equation}
where $\bm{\hat{F}}_A$, $\bm{\hat{F}}_V \in \mathbb{R}^{T \times d}$ denote the updated temporal features for audio and visual modalities, $\bm{W}^Q,\bm{W}^K,\bm{W}^V \in \mathbb{R}^{d \times d}$ are learnable parameters, and $\delta$ is the softmax function.

During this process, the modality-specific features $\bm{F}_A$ and $\bm{F}_V$ query the anchor features $\bm{Z}_{av}$ to assess whether the event categories encoded by $\bm{Z}_{av}$ are present at each timestamp, reflected by higher cross-attention weights. 
In this way, salient event semantics are propagated from the anchor features to temporally aligned positions, facilitating fine-grained, timestamp-level prediction.

\subsection{Classification and Model Training}

Inspired by prior work on temporal action localization~\cite{CO2Net, Li_2023_CVPR}, we introduce modality-specific classifiers to predict the foreground weight for each audio and visual modality, with $\bm{w}_a, \bm{w}_v \in \mathbb{R}^{T \times 1}$. 
Foreground segments with high probability indicate that events, rather than background noise, are more likely to occur.
Specifically, the classifier is applied to the final audio or visual features ($\bm{\hat{F}_A}$/$\bm{\hat{F}_V}$), implemented by a linear layer followed by a sigmoid activation. 
We use the average of $\bm{w}_a$ and $\bm{w}_v$ to represent the overall event foreground score, denoted as $\bm{\overline{w}} \in \mathbb{R}^{T \times 1}$.

Meanwhile, the audio-visual event probability $\bm{p}_{av} \in \mathbb{R}^{T \times C}$ can be obtained via an additional classifier applied to the concatenated audio and visual features, \ie, $[\bm{\hat{F}_A}; \bm{\hat{F}_V}]$. 
Then, $\bm{p}_{av}$ is modulated by the event foreground score $\bm{\overline{w}}$ to suppress background temporal segments, computed as $\bm{p}_{av} \odot \bm{\overline{w}}$, where $\odot$ denotes element-wise multiplication.

The modulated $\bm{p}_{av}$ can be used for temporal-level prediction during the inference phase. 
For model training, we compute the video-level audio-visual event prediction $\bm{\hat{p}} \in \mathbb{R}^{1 \times C}$ by aggregating $\bm{p}_{av} \in \mathbb{R}^{T \times C}$ over the temporal dimension using MIL pooling, as in~\cite{tian2018audio}.
Then, the cross-entropy loss between $\bm{\hat{p}}$ and the video-level ground truth label $\bm{y}$ is computed.
Notably, $\bm{P}_a$ and $\bm{P}_v$ used in MEAE module (Eq.~\ref{eq:Pa_Pv}) are regularized in the same way.

\section{Experiments}
\subsection{Experimental Setups}
\noindent\textbf{Datasets.}
We evaluate our method on the UnAV-100~\cite{geng2023dense} and ActivityNet1.3~\cite{heilbron2015activitynet} datasets.
\textit{UnAV-100} serves as the standard benchmark for the fully-supervised DAVEL task. 
It contains 10,790 untrimmed videos totaling over 126 hours of video. 
The videos have varying durations, with most exceeding 40 seconds. 
The video content spans 100 common audio-visual event categories drawn from real-world scenarios, including human and animal activities, musical instrument performances, and various means of transportation, \etc. 
Each video contains an average of 2.8 audio-visual events, with a maximum of 23 events per video. 
According to the standard dataset division scheme, the training, validation, and test sets follow a 3:1:1 split.
\textit{ActivityNet1.3} is another large-scale dataset, originally constructed for temporal action localization. 
We also evaluate on this dataset to assess the robustness of our approach. 
Specifically, it includes 19,994 long video clips, each with an average duration of approximately 100 seconds and 1.41 activity classes. 
This dataset covers 203 activity categories, including daily activities, sports competitions, and performing arts.

\noindent\textbf{Evaluation Metric.}
Following prior work~\cite{geng2023dense}, we adopt mean Average Precision (mAP) as the primary evaluation metric to assess temporal localization performance.
We report mAP values at tIoU thresholds ranging from 0.5 to 0.9 with a step size of 0.1 ([0.5:0.1:0.9]). 
Meanwhile, the average mAP across this threshold range (`Avg.') is reported as a holistic measure of overall model performance.

\noindent\textbf{Implementation Details.}
\textit{(1) Feature extraction.} We preprocess the audio and visual data following prior works~\cite{geng2023dense,heilbron2015activitynet}.
For the UnAV-100 dataset, we sample the video frames at 25 FPS. Then, the RAFT~\cite{teed2020raft} model is employed to extract the optical flow images. Then, they are combined with 24 consecutive RGB frames and sent to a two-stream pretrained I3D~\cite{carreira2017quo} model to extract 2048-D visual features.
The audio signal is segmented at time intervals of 0.96 seconds using a sliding window of 0.32 seconds. The pretrained VGGish~\cite{hershey2017vggish} is used to extract 128-D audio features.
To facilitate parallel processing of videos with varying duration, a maximum temporal length of $T$=224 is set by clipping or zero-padding.
For the ActivityNet1.3 dataset, the sampling rates of video frames and the audio stream are 16 FPS and 16 KHz, respectively. The audio and visual features are extracted using the pretrained VGGish and R(2+1)D-34 model, respectively. $T$ is set to 256.
\textit{2) Model Configuration.} 
For model training on the UnAV-100 dataset, we train our model for 40 epochs with a batch size of 16. For the training on ActivityNet1.3, the total training epoch and the batch size are 40 and 8, respectively.
The Adam optimizer is utilized with a learning rate of 1e-4 for both datasets.
{The optimal configuration of the key hyperparameters, such as the number of global anchors $K$ and the number of anchors in each local window $k$, are set to 10 and 4, respectively. The number of local temporal windows $M$ is 14 and 16 for the UnAV-100 and ActivityNet1.3 datasets, respectively.}
More detailed analysis on these parameters will be provided in the ablation study.
During the inference phase, we utilize the conventional threshold of 0.5 to identify confident event classes from the prediction $\bm{p}_{av}$.
All the experiments are conducted on a NVIDIA A40 (40GB) GPU. Codes will be released.

\begin{table}[t]
  \centering
  \Huge
      \resizebox{\linewidth}{!}{
  \begin{tabular}{l|c|ccccc|c}
    \toprule
    Methods    & Task & 0.5 & 0.6 &0.7 &0.8 &0.9 &Avg.\\ \midrule
    AVEL~\cite{tian2018audio}  & \multirow{4}{*}{AVEP} &  18.7 & 16.5 & 14.5 & 12.3 & 10.0 & 19.2 \\
    JoMoLD~\cite{cheng2022joint}  & & 22.4 & 19.6 & 17.3 & 14.8 & 12.0 & 22.9 \\
    PPAE~\cite{gao2025learning} &   & 23.6 & 20.5 &17.9 &15.2 &12.4 &23.8\\
    NREP~\cite{jiang2024resisting}  &  & 24.9 &19.7 &17.2 &15.6 & \underline{15.2} &26.4 \\ \midrule
    DAVE~\cite{geng2023dense} & \multirow{2}{*}{DAVEL} &21.6 &19.5 &17.1 &14.8 &11.7 &21.6 \\ 
    CCNet~\cite{zhou2025dense}  &  & \underline{27.0} & \underline{24.4} & \underline{21.3} & \underline{17.8} & 13.6 & \underline{26.9} \\ \midrule
    \textbf{CLASP (ours)}  & DAVEL & \textbf{29.6} & \textbf{26.3} & \textbf{23.0} & \textbf{19.7} & \textbf{15.8} & \textbf{30.0} \\ 
  \bottomrule
\end{tabular}
}
  \caption{\textbf{Comparison with prior works on UnAV-100 dataset.} 
  All methods are trained and evaluated under weakly supervised settings. The best and second-best results are highlighted in bold and underlined, respectively.}
  \label{tab:sota_comparsion_UNAV}
\end{table}

\begin{table}[t]
\Huge
      \resizebox{\linewidth}{!}{
  \begin{tabular}{c|ccccc|c}
    \toprule
    Methods   &0.5 & 0.6 &0.7 &0.8 &0.9 &Avg.\\ \midrule
    {ActionFormer~\cite{actionformer}}  &23.2 &19.7 &16.2 &11.8 &6.2 &14.3\\ 
    DAVE~\cite{geng2023dense}  &25.5 &21.7 &17.8 &13.0 &6.8 &15.8\\ 
    CCNet~\cite{zhou2025dense}  &\underline{31.7} &\underline{27.0} &\underline{22.2} &\underline{16.3} &\underline{8.5} &\underline{19.6}\\  \midrule
    \textbf{CLASP (ours)} & \textbf{34.6} & \textbf{29.0} & \textbf{23.6} & \textbf{17.6} & \textbf{9.8} & \textbf{21.3} \\ 
  \bottomrule
\end{tabular}
}
  \caption{{\textbf{{Comparison on the ActivityNet1.3 dataset.}}}
  }
  \label{tab:sota_comparsion_act}
\end{table}

\subsection{Main Comparison Results} 

We evaluate our method on the UnAV-100 and ActivityNet1.3 datasets and compare it against prior state-of-the-art (SOTA) methods from related tasks.

\noindent\textbf{Evaluation on the UnAV-100 dataset.}
Table~\ref{tab:sota_comparsion_UNAV} presents the evaluation results.
We compare with SOTA methods designed for the audio-visual event perception (AVEP) task, including AVEL~\cite{tian2018audio}, JoMoLD~\cite{cheng2020look}, PPAE~\cite{gao2025learning}, and NREP~\cite{jiang2024resisting}.
As shown in the table, our method outperforms all these approaches by a significant margin.
For instance, our method surpasses the best method, NREP, by 3.6\% on the `Avg.' metric.
In addition, our method also outperforms SOTA approaches developed for the DAVEL task.
The baseline model DAVE~\cite{geng2023dense} achieves only 21.6\% on the `Avg.' metric.
The recent SOTA method for fully-supervised DAVEL, CCNet~\cite{zhou2025dense}, achieves a higher score of 26.9\%.
However, they still fall significantly short of our method under all tIoU metrics.
Notably, we do not compare with LoCo~\cite{xing2024loco}, which performs comparably to CCNet on the fully-supervised DAVEL task, as its implementation is not publicly available so far.
It is worth noting that fully-supervised DAVEL methods can reach around 50\% `Avg.'. This performance gap highlights the difficulty and potential of the studied W-DAVEL task.

\noindent\textbf{Evaluation on the ActivityNet1.3 dataset.}
Table~\ref{tab:sota_comparsion_act} presents the evaluation results on ActivityNet1.3 dataset.
Our method again outperforms prior SOTA methods from related tasks, surpassing the most competitive method, CCNet~\cite{zhou2025dense}, by 1.7\% on the `Avg.' metric.
These results demonstrate the effectiveness and generalizability of our method.

\subsection{Ablation Studies}
In this section, we conduct extensive experiments to evaluate the impacts of some key hyperparameters and the effectiveness of each main module design.
The ablation experiments are conducted on the standard UnAV-100 dataset.

\noindent\textbf{Impact of the Global Anchor Number $K$.} In our Global Anchor Identification mechanism, we identify $K$ anchor features from the entire video.
We conduct experiments to evaluate its impact by varying the value of $K$.
The results are shown in Table~\ref{tab:ablation_K}.
When using quite a few anchors, \eg, $K=5$, the model has a lower performance at 29.4\% on `Avg.' metric.
Increasing the global anchor number to 10, the model's performance improves accordingly.
However, continuing to enlarge the number of $K$ introduces a slight decrease in performance.
Therefore, we adopt $K=10$ as the optimal setup.

\noindent\textbf{Impact of the Local Anchor Number $k$.}
In our Local Anchor Identification mechanism, we identify $k$ anchor features from each local temporal window.
We conduct experiments to evaluate its impact.
As shown in Table~\ref{tab:ablation_k}, the optimal performance is achieved when $k=4$.
Similar to the trend observed for the aforementioned $K$, the model performance gradually improves with increasing $k$ before declining when $k$ exceeds 4.
Therefore, we set $k=4$ as the default setting.

\begin{table}[t]
\centering
  \setlength{\tabcolsep}{3mm}
  \begin{tabular}{c|ccccc|c}
    \toprule
    $K$  &0.5 & 0.6 &0.7 &0.8 &0.9 &Avg.\\ \midrule
    5  & 29.1 & 25.6 &22.4 &19.1 &15.2 &29.4\\
    \textbf{10} & \textbf{29.6} & \textbf{26.3} & \textbf{23.0} & \textbf{19.7} & \textbf{15.8} & \textbf{30.0} \\    
    15 & {29.3} &{25.9} &{22.9} &{19.4} &15.1 &{29.6}\\
    20 & 28.4 &25.4 &22.1 &19.1 &15.8 &29.1\\
  \bottomrule
\end{tabular}
  \caption{\textbf{Impact of the global anchor number $K$.}}
  \label{tab:ablation_K}
\end{table}

\begin{table}[t]
\centering
  \setlength{\tabcolsep}{3mm}

  \begin{tabular}{c|ccccc|c}
    \toprule
    $k$  &0.5 & 0.6 &0.7 &0.8 &0.9 &Avg.\\ \midrule
    1  & 28.3 & 25.4 &22.1 &19.2 &\underline{15.7} &29.2\\
    2 & \textbf{30.4} &\textbf{27.1} &\textbf{23.3} &\underline{19.4} &14.3 &\underline{29.4}\\
    \textbf{4} & \underline{29.6} & \underline{26.3} & \underline{23.0} & \textbf{19.7} & \textbf{15.8} & \textbf{30.0} \\    
    8 & 28.9 &25.8 &22.5 &19.1 &15.4 &29.3\\
  \bottomrule
\end{tabular}
  \caption{\textbf{Impact of the local anchor number $k$.}}
  \label{tab:ablation_k}
\end{table}

\begin{table}[t]
\centering
  \setlength{\tabcolsep}{3mm}

  \begin{tabular}{c|ccccc|c}
    \toprule
    $M$  &0.5 & 0.6 &0.7 &0.8 &0.9 &Avg.\\ \midrule
    4 & 29.1 & 25.8 & 22.5 & 19.3 &15.3 & 29.4\\
    8 & 29.4 & 26.0 & 22.6 & 19.2 &15.1 & 29.7\\
    \textbf{14} & \textbf{29.6} & \textbf{26.3} & \textbf{23.0} & \textbf{19.7} & \textbf{15.8} & \textbf{30.0} \\    
    28 & 29.4 &25.9 &22.5 &19.2 &15.1 &29.6\\
  \bottomrule
\end{tabular}
  \caption{  \textbf{Impact of the local temporal window number $M$.} }
  \label{tab:ablation_M}
\end{table}

\begin{figure*}[t]
    \centering
  \includegraphics[width=\linewidth]{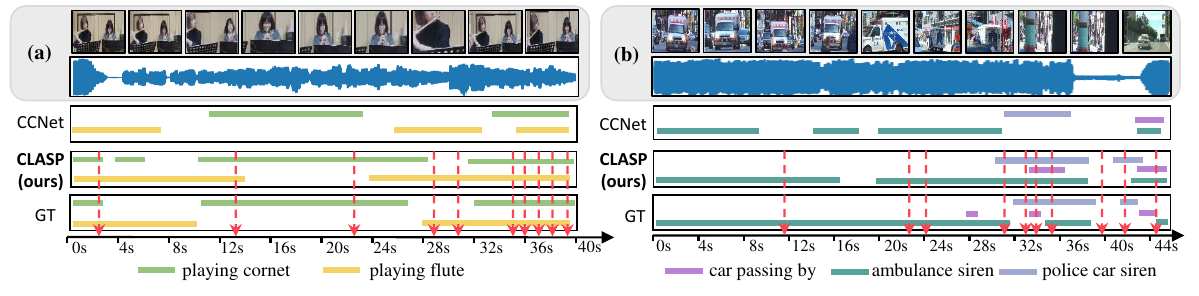}
    
    \caption{\textbf{Visualization examples of the weakly-supervised dense audio-visual event localization.} We compare our method with CCNet~\cite{zhou2025dense}. `GT' is the ground truth. The red arrows mark the identified global ten salient anchor timestamps.
    }
    \label{fig:visualization_results}
\end{figure*}

\noindent\textbf{Impact of the Local Temporal Window Number $M$}.
In our Local Anchor Identification mechanism, we divide the entire video into $M$ temporal windows and identify the $k$ salient anchors in each window.
We fix $k$ to 4 and conduct experiments to explore the impact of $M$.
As shown in Table~\ref{tab:ablation_M}, the optimal performance is achieved when $M=14$.
Using fewer or more temporal windows, the model's performance slightly decreases.
Therefore, we employ $M=14$ as the optimal setup.
It is noteworthy that our model still maintains better performance compared to prior approaches under varied hyperparameters ($K$, $k$, and $M$), demonstrating the superiority and robustness of our framework.

\noindent\textbf{Ablation Study on the MEAE Module.}
In our Mutual Event Agreement Evaluation (MEAE) module, we utilize the initial audio and visual features ($\bm{a}, \bm{v}$) extracted by pretrained backbones to compute the agreement score.
We also explore a variant that employs the features after going through the unimodal transformers ($\bm{F}_A, \bm{F}_V$).
The experimental results are shown in Table~\ref{tab:ablation_MEAE}.
Using the vanilla audiovisual features $\bm{a}$ and $\bm{v}$ has superior performance.
This indicates that it is not necessary to perform unimodal temporal relation modeling in the mutual event agreement evaluation phase.

\begin{table}[t]
\centering
  \setlength{\tabcolsep}{2.5mm}

  \begin{tabular}{c|ccccc|c}
    \toprule
    Setups   &0.5 & 0.6 &0.7 &0.8 &0.9 &Avg.\\ \midrule
    $\bm{F}_A,\bm{F}_V$  &29.0 &25.9 &22.8 &19.5 &15.3 &29.6\\ 
    $\bm{a}, \bm{v}$  &\textbf{29.6} & \textbf{26.3} & \textbf{23.0} & \textbf{19.7} & \textbf{15.8} & \textbf{30.0} \\  
  \bottomrule
\end{tabular}

  \caption{\textbf{Ablation study on MEAE module.} We explore the impacts of different audio and visual features used in mutual event agreement score computation.}
  \label{tab:ablation_MEAE}
\end{table}

\begin{table}[t]
\centering
  \begin{tabular}{cc|ccccc|c}
    \toprule
    GAI &LAI   &0.5 & 0.6 &0.7 &0.8 &0.9 &Avg.\\ \midrule
    \ding{56} &\ding{52} & 29.4 &26.2 &22.3 &19.4 &15.5 &29.8\\
    \ding{52} &\ding{56}  &29.2 &25.9 &22.8 &19.3 &15.1 &29.5\\ 
    \ding{52} &\ding{52}  & \textbf{29.6} & \textbf{26.3} & \textbf{23.0} & \textbf{19.7} & \textbf{15.8} & \textbf{30.0} \\ 
  \bottomrule
\end{tabular}
  \caption{\textbf{Ablation study on the two anchor identification mechanisms in CSAI module.} `GAI' and `LAI' are abbreviations for Global Anchor Identification and Local Anchor Identification mechanisms in CSAI.}
  \label{tab:ablation_CSAI_mechanism}
\end{table}

\noindent\textbf{Ablation Study on the CSAI Module.}
Our Cross-modal Salient Anchor Identification (CSAI) module introduces two mechanisms: Global Anchor Identification (GAI) and Local Anchor Identification (LAI).
We investigate their contributions by ablating each mechanism.
As shown in Table~\ref{tab:ablation_CSAI_mechanism}, removing either the GAI or LAI mechanism leads to a performance drop.
These results highlight the effectiveness of both mechanisms and suggest their complementarity.
We also observe that using only LAI yields slightly better performance than using only GAI.
This indicates that anchor identification within multiple local temporal windows is more beneficial, as it facilitates finer-grained semantic comparisons over time.

\noindent\textbf{Ablation Study on the ATP Module.}
Our Anchor-based Temporal Propagation (ATP) module utilizes the identified multimodal anchor features to interact with the temporal audio and visual features. 
We conduct ablation experiments to explore the impact of anchor features.
As shown in Table~\ref{tab:ablation_ATP}, without using the anchors features in ATP, \ie, only using the modality-specific audio and visual features obtained by unimodal transformers, the model's performance drops in all metrics, resulting in an overall decrease of 0.7\% on `Avg.' metric.
These results demonstrate the benefits of anchor features. The contained salient semantics of the audio-visual events in the anchor features can be propagated to enhance the temporal features, facilitating the event localization under the weakly supervised setting.

\noindent\textbf{Ablation Study on the Modality.}
Our CSAI module identifies anchors for both audio and visual modalities (Eq.~\ref{eq:anchor_a_v}), which are then used in the ATP module (Eq.~\ref{eq:ATP}).
We explore the impact of the anchor utilization from a single modality.
The experimental results are shown in Table~\ref{tab:ablation_CSAI_modality}.
Compared to the joint utilization of anchors from both audio and visual modalities, using only the anchor features from a single modality, the model's average performance `Avg.' drops by 0.4\% and 0.6\%, respectively. 
This is reasonable since the localization of target audio-visual events requires a joint understanding of both audio and visual modalities. The anchor features aggregated from both modalities are beneficial for event semantic propagation along timeline.

\begin{table}[t]
\centering
  \begin{tabular}{c|ccccc|c}
    \toprule
    Setups   &0.5 & 0.6 &0.7 &0.8 &0.9 &Avg.\\ \midrule
    w/o. anchor  &28.9 &25.6 &22.5 &19.3 &15.2 &29.3\\ 
    w. anchor & \textbf{29.6} & \textbf{26.3} & \textbf{23.0} & \textbf{19.7} & \textbf{15.8} & \textbf{30.0} \\ 
  \bottomrule
\end{tabular}
  \caption{\textbf{Ablation study on ATP module.} `w/o. anchor' represents that the identified anchor features are not used.}
  \label{tab:ablation_ATP}
\end{table}

\begin{table}[t]
\centering
      \resizebox{\linewidth}{!}{
  \begin{tabular}{c|ccccc|c}
    \toprule
    Modality   &0.5 & 0.6 &0.7 &0.8 &0.9 &Avg.\\ \midrule
    Audio  &29.2 &25.8 &22.5 &19.5 &\underline{15.5} & \underline{29.6} \\ 
    Visual  & \textbf{30.4} &\textbf{27.1} &\textbf{23.3} &19.4 &14.3 & 29.4\\
    \textbf{Audio + Visual} & \underline{29.6} & \underline{26.3} & \underline{23.0} & \textbf{19.7} & \textbf{15.8} & \textbf{30.0} \\   
  \bottomrule
\end{tabular}
}
  \caption{\textbf{Ablation study on the modality used for anchor identification and temporal propagation.}}
  \label{tab:ablation_CSAI_modality}
\end{table}

\subsection{Qualitative Results}
We present qualitative visualizations for dense audio-visual event localization and compare our method with the prior SOTA method CCNet~\cite{zhou2025dense}. 
As illustrated in Fig.~\ref{fig:visualization_results}(a), the video contains two audio-visual events in a musical performance scene: \textit{playing cornet} and \textit{playing flute}.
Both CCNet and our method correctly identify these two events. However, CCNet fails to localize the \textit{playing cornet} during the initial two seconds. 
Moreover, our method provides more accurate temporal boundaries for both events between 28s and 40s.
Similarly, in Fig.~\ref{fig:visualization_results}(b), CCNet misses the \textit{car passing by} and \textit{ambulance siren} events occurring between 32s and 40s, while our method successfully identifies both.
We also mark the positions of the identified global salient anchors ($K$=10) using red arrows.
These anchors are unevenly distributed, often aligning with the moments where major events occur. 
The identified anchors play a crucial role in guiding event semantic perception and propagating it to the remaining temporal timestamps through our anchor-based mechanism, thereby enhancing temporal localization under weak supervision.

\section{Conclusion}
We explore the W-DAVEL task, aiming to achieve dense audio-visual event localization under a weakly-supervised setting where only the video-level event labels are provided.
Our method, CLASP, addresses this task by exploiting the cross-modal salient anchors, \ie, temporal timestamps that exhibit the most consistent event semantics across audio and visual modalities.
A mutual event agreement evaluation module is used to assess the consistency of audio and visual event predictions.
The resulting agreement score serves as the indicator for identifying cross-modal salient anchors through both global video-level and local temporal window-level comparisons.
The identified anchor features are then leveraged to enhance temporal audio and visual features, which are used for final event classification.
Extensive experiments verify the effectiveness of our method.
We hope our work encourages further research into this challenging yet practical task.

\bibliography{aaai2026}


\end{document}